\documentclass[11pt]{article}

\usepackage[preprint]{acl}

\usepackage{times}
\usepackage{latexsym}

\usepackage[T1]{fontenc}
\usepackage[utf8]{inputenc}

\usepackage{microtype}

\usepackage{inconsolata}

\usepackage{graphicx}

\usepackage{booktabs}
\usepackage{amsmath}
\usepackage{url}

\title{EpiCurveBench: Evaluating VLMs on Epidemic Curve Digitization}

\author{
  Thomas Berkane\thanks{\ Corresponding author: \texttt{thomas.berkane@childrens.harvard.edu}} \quad Maimuna S.\ Majumder \\
  Computational Health Informatics Program \\
  Boston Children's Hospital \& Harvard Medical School \\
}

\begin{document}
\maketitle
\begin{abstract}
Chart-to-data extraction with vision-language models (VLMs) is increasingly evaluated on benchmarks that show diminishing headroom (frontier VLMs exceed 89\% on ChartQA) and with metrics that treat extracted points as unordered key-value pairs, ignoring the temporal structure of time series and penalizing small alignment shifts as catastrophic failures. We address both gaps with EpiCurveBench, a benchmark of 1{,}000 real-world epidemic curve images curated from diverse public-health sources, and EpiCurveSimilarity (ECS), an evaluation metric that aligns predicted and ground-truth series via dynamic programming, tolerating local temporal shifts and gaps while penalizing them proportionally. Evaluating six methods---three frontier closed VLMs, one open VLM, and two specialized chart-extraction systems---we find the strongest model reaches only 52.3\% ECS, and that ECS spreads the four general-purpose VLMs over a 25-point range where key-value metrics (RMS, SCRM) compress them into a 5-point band. We further validate ECS against four downstream epidemiological summary statistics, finding that higher ECS predicts smaller errors in total counts, peak timing, and peak magnitude, and higher growth-rate fidelity; across all four, ECS correlates 1.5--3.6$\times$ more strongly than Dynamic Time Warping, which lacks a gap penalty and therefore cannot distinguish a truncated prediction from a temporally faithful one. EpiCurveBench targets a high-impact public-health application---unlocking decades of outbreak data trapped in published figures---but the benchmark and metric apply directly to any structured time-series chart-extraction setting.
\end{abstract}

\begin{figure}[t]
\centering
\includegraphics[width=\columnwidth]{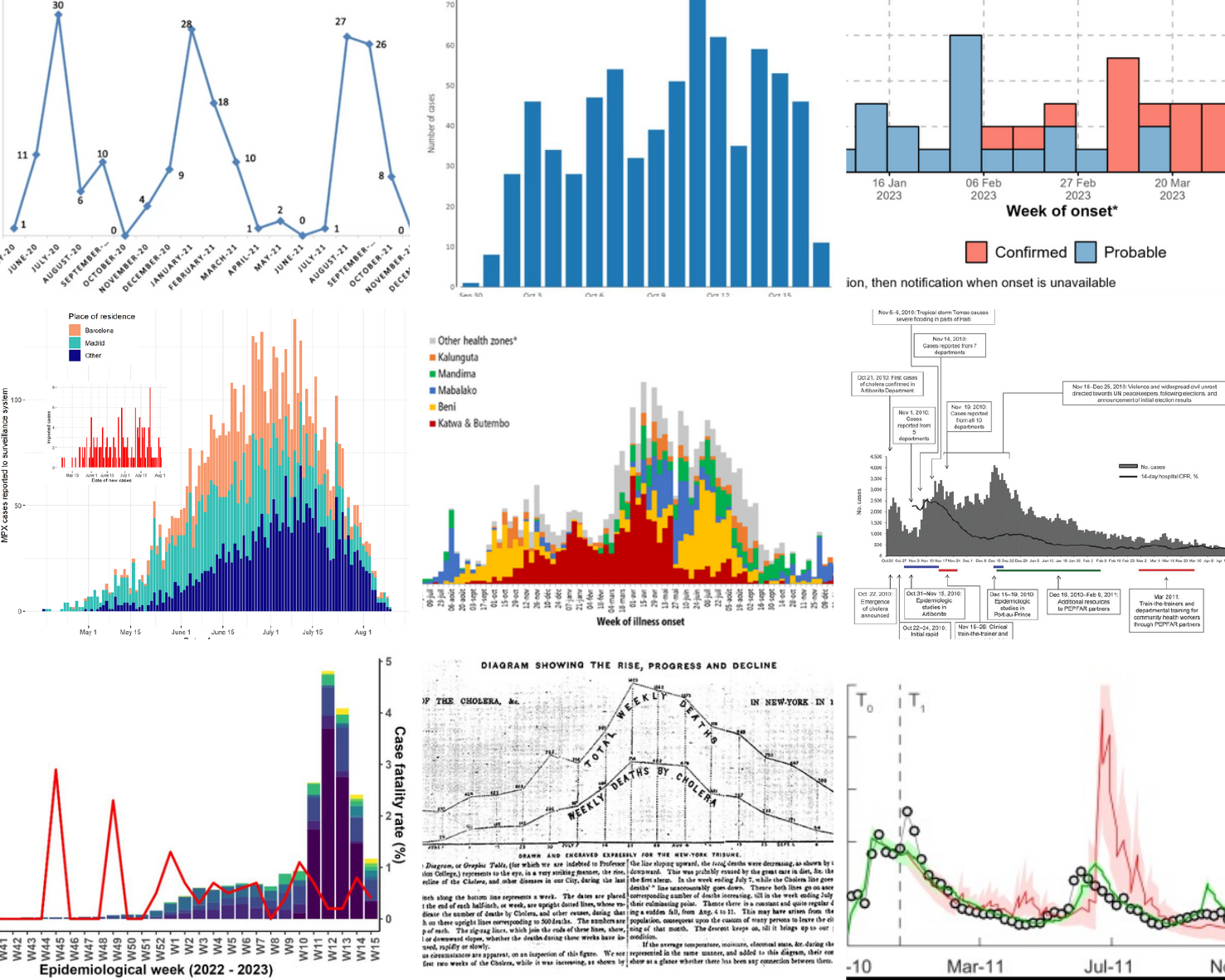}
\caption{Sample images from EpiCurveBench. Parts of the images, including axes, are truncated for space.}
\label{fig:epicurve_examples}
\end{figure}

\section{Introduction}
Vision-language models (VLMs) are increasingly used to extract structured numerical data from chart images, but the benchmarks and metrics used to track progress on this task are showing their limits. Public benchmarks such as ChartQA~\citep{chartqa} are dominated by simple, mostly synthetic infographics with sparse datapoints and clearly printed values, and frontier VLMs now exceed 89\% Relative Mapping Similarity (RMS) on them (Appendix~\ref{app:benchmark_comparison}). At the same time, RMS---the most widely used evaluation metric---treats extracted points as unordered key-value pairs and penalizes any temporal shift as a complete mismatch, conflating value-correct-but-shifted extractions with completely wrong ones (Figure~\ref{fig:metric_comparison}). The community lacks (i) a real-world stress-test benchmark for dense time-series chart extraction, and (ii) a metric that scores such extractions in a way that reflects their downstream usefulness.

We instantiate both contributions in a high-stakes domain: epidemic curves (``epicurves''), the time-series chart format in which public-health agencies publish case, hospitalization, and death counts during outbreaks. Much of this data is published only as images embedded in PDF reports, so improving automated chart extraction directly unlocks training data for the forecasting models that guide outbreak response~\citep{forecast}. Compared to the charts in existing benchmarks, epicurves are markedly harder: they contain dense datapoints (median 156 datapoints per chart, up to 8{,}592), overlapping or stacked series, rotated or tiny axis text, low-resolution scans of historical reports, thin bars, low contrast, and overlaid annotations. Human digitization of a single complex epicurve can take up to three hours.

We introduce \textbf{EpiCurveBench}, a benchmark of 1{,}000 real-world, openly licensed epicurve images spanning 14 diseases, 37 countries, and the years 1849--2025. It pairs 900 charts with source-aligned tabular ground truth (audited for alignment on a 20\% sample) with 100 manually annotated charts curated via a three-stage protocol that maximizes stylistic diversity and verifies non-redundancy with DINOv2~\citep{dinov2} embeddings.

We introduce \textbf{EpiCurveSimilarity (ECS)}, an evaluation metric based on Edit Distance with Real Penalty~\citep{erp}. ECS aligns predicted and ground-truth series via dynamic programming, tolerates local temporal shifts and gaps, and normalizes substitution penalties by each series' own y-axis range. While developed for epicurves, ECS depends only on the structure of the underlying task (ordered time-indexed series with sparse labels) and applies directly to other structured time-series chart-extraction settings.

We evaluate six state-of-the-art methods---three frontier closed VLMs, one open VLM, and two specialized chart-extraction systems---on EpiCurveBench. The strongest model reaches only 52.3\% ECS. ECS spreads the four general-purpose VLMs over a 25-point range where the key-value metrics RMS and SCRM compress them into a $\sim$5-point band, and per-series ECS correlates 1.5--3.6$\times$ more strongly than DTW with downstream errors in total counts, peak timing, peak magnitude, and growth-rate fidelity.

\textbf{Contributions.} (1) \textit{Benchmark}: EpiCurveBench, a challenging real-world testbed for dense time-series chart extraction. (2) \textit{Metric}: EpiCurveSimilarity, a temporally-aware metric that discriminates between general-purpose VLMs where RMS and SCRM do not, and correlates more strongly than DTW with four downstream epidemiological statistics. (3) \textit{Empirical findings}: a comprehensive evaluation showing that even leading general-purpose VLMs with reasoning and tool-use fall well short of usable extraction quality on real-world public-health charts. (4) \textit{Artifacts}: open-source data\footnote{\url{https://huggingface.co/datasets/tberkane/EpiCurveBench}} and evaluation pipeline.\footnote{\url{https://github.com/tberkane/EpiCurveBench}}

\section{Background and Related Work}
\label{sec:related_work}

\paragraph{Chart data extraction.} Early pipelines combined classical computer vision with OCR~\citep{cv} but were brittle to stylistic variation. VLM-based approaches---MatCha~\citep{matcha}, DePlot~\citep{deplot}, and specialized systems such as OneChart~\citep{onechart} and TinyChart~\citep{tinychart}---enable end-to-end extraction without hand-engineered pipelines, and frontier general-purpose VLMs (GPT-5.2~\citep{gpt52}, Claude Opus 4.5~\citep{opus45}, Gemini 2.5 Pro~\citep{gemini25pro}) show strong zero-shot performance but still struggle on complex real-world charts~\citep{plottwist}.

\paragraph{Chart benchmarks.} Existing data-extraction benchmarks (PlotQA~\citep{plotqa}, ChartQA~\citep{chartqa}) are dominated by simple, mostly synthetic infographics with few datapoints; frontier VLMs exceed 89\% RMS on ChartQA (Appendix~\ref{app:benchmark_comparison}). Chart-QA benchmarks like ChartQAPro~\citep{chartqapro} and CharXiv~\citep{charxiv} broaden chart diversity but evaluate natural-language answers, not structured data. General multimodal-reasoning benchmarks (MMMU~\citep{mmmu}, MathVista~\citep{mathvista}) include charts only as one of many image categories, with multiple-choice answers.

\paragraph{Evaluation metrics.} The most widely used metric, \textbf{Relative Mapping Similarity (RMS)}~\citep{deplot}, treats extracted data as unordered key-value mappings with composite keys (series name $+$ x-axis label), matched by textual similarity. \textbf{SCRM}~\citep{scrm} adds chart-type-aware tolerances but follows the same key-value paradigm. Both are insensitive to order and penalize any x-axis label mismatch as a complete miss, conflating wrong values with right-but-shifted ones (Figure~\ref{fig:metric_comparison}). Time-series similarity measures avoid these issues: \textbf{Dynamic Time Warping (DTW)}~\citep{dtw} enables monotone alignment by allowing one-to-many matches but lacks an explicit gap penalty for unmatched points, while \textbf{Edit Distance with Real Penalty (ERP)}~\citep{erp} treats insertions and deletions as first-class operations with configurable real-valued penalties. ECS extends ERP with per-series y-axis normalization, a tolerance threshold, and label-similarity-based series matching. Section~\ref{sec:metric_comparison} compares ECS empirically against both RMS and SCRM on extraction-quality discrimination; Section~\ref{sec:downstream} compares ECS against DTW on correlation with downstream summary statistics.

\section{EpiCurveBench}
\subsection{Data Collection}
EpiCurveBench contains 1{,}000 real-world, openly licensed epidemic curve (epicurve) images from two complementary sources: (i) 900 epicurves whose original sources provide the underlying numerical data in tabular format (referred to as Set 1), and (ii) 100 epicurves that we manually annotate to broaden stylistic diversity and validate performance under realistic conditions where no accompanying tabular data is available (referred to as Set 2). The dataset spans 13 source types (Appendix~\ref{app:dataset_distributions}).

To enable large-scale quantitative evaluation without prohibitively time-consuming manual labeling, we collect 900 epicurve images whose sources also provide corresponding case data in machine-readable form (e.g., a table in the same report or an accompanying downloadable dataset). In practice, sources are highly constrained: in public health reporting, it is uncommon for organizations to publish both an epicurve figure and the underlying time series used to generate it. As a result, most of these 900 epicurves come from the US Centers for Disease Control and Prevention (CDC), one of the few entities that provides epidemic curves with retrievable ground-truth case counts. We supplement the CDC-derived set with additional sources that expose ground truth to increase visual and formatting variety, but such sources are rare and contribute comparatively few curves.

To test generalization beyond the subset of charts that happen to be published with attached data, we additionally collect a set of 100 openly licensed epicurves selected for maximal stylistic variety. We collect these images by crawling directory pages and linked outbreak report PDFs from organizations such as WHO regional offices, PAHO, and the CDC, and by running targeted Google Image searches to capture additional formats not well represented in ground-truth-backed sources. From an initial pool of 1{,}397 candidate images, we apply a three-stage curation protocol. First, an annotator incrementally narrows the pool by selecting charts that maximize visual dissimilarity across source, disease, chart type, and stylistic features (e.g., dense datapoints, overlapping or stacked series, rotated or tiny text, low resolution, thin bars, overlaid annotations), yielding 100 retained charts. Second, we embed the retained images with DINOv2~\citep{dinov2} (ViT-L/14) and apply DBSCAN clustering on cosine distances to flag near-duplicates (cosine similarity $>0.9$); 8 of the 100 charts were flagged as redundant. Third, the annotator draws replacements for flagged charts from the remaining candidate pool and verifies that each is visually distinct from the rest of the set. The final Set 2 spans 13 source types, 14 diseases, 36 countries, and 3 chart types.

To validate that the 900 tabular-derived charts (Set 1) are correctly paired with their source tables, a human annotator audited a random sample of 180 charts (20\% of Set 1), comparing each image against its tabular ground truth. All 180 charts (100\%) were rated ``Aligned,'' supporting use of the source tables as ground truth without further per-chart re-annotation.

\subsection{Data Annotation}
We manually annotate Set 2 using WebPlotDigitizer~\citep{WebPlotDigitizer}; the remaining 900 epicurves already include ground-truth time series from their original sources. For each manually annotated image, we calibrate the x- and y-axes using two known reference points per axis, then click each visible datapoint to record its x and y values. Each epicurve produces a CSV containing the full set of x-axis values and one or more series columns, preserving the original axis labels and series names from the source figure. When multiple series are present, we extract all visually distinct, labeled series, including non--case-count quantities when plotted (e.g., deaths or rainfall).
To ensure consistent handling of ambiguous cases, our protocol follows three principles. First, we annotate only data that is explicitly visible and clearly indicated in the figure, making no assumptions about missing, truncated, or implied values. Second, for stacked bar charts, we treat each stack segment as a separate time series when the figure provides distinct segment labels (via a legend or direct on-chart labeling). Third, when stacked segments are unlabeled, incompletely labeled, or visually indistinguishable, we annotate only the aggregate total (i.e., the full stack height) rather than inferring component series. We provide the same instructions to evaluated models to align human and automated outputs.

Each curve was independently annotated by two annotators, yielding a mean inter-annotator ECS of 90.9\% (median 95.4\%; full stratification by chart type and series length in Appendix~\ref{app:annotation_agreement}). Disagreements almost exclusively involve surplus or missed datapoints at series boundaries (e.g., one annotator including an additional leading or trailing datapoint that is visually ambiguous); annotators never disagreed on series identity.

For charts mixing bar and line visualizations on different y-axes (n=129, 12.9\% of the benchmark), our annotation protocol stores all series as columns sharing a single time axis, regardless of their rendering type. The extraction prompt given to models (Section~\ref{sec:methods}) is correspondingly visualization-agnostic. During evaluation, ECS scores each matched series independently and normalizes by that series' own y-axis range (Section~\ref{sec:metric}), so a bar series in absolute counts and a line series in percent are each scored on their own scale.

\subsection{Dataset Characteristics}
\label{sec:dataset_characteristics}
Figure~\ref{fig:epicurve_examples} shows sample epicurves from EpiCurveBench, illustrating the benchmark's wide range of visual styles and formats. Common challenges include thin bars, low resolution, text overlays, stacked bars with similar colors, rotated or tiny axis text, and historical plots with partially illegible values.

Table~\ref{tab:dataset_stats} summarizes the per-chart statistics of EpiCurveBench. The benchmark contains charts of widely varying complexity: while the median chart has 2 series and 156 datapoints, the long tail extends to charts with 14 series and over 8{,}000 datapoints. Chart-type distribution is line 71.4\%, bar 15.7\%, both 12.9\%; cumulative vs.\ non-cumulative is 13.6\% vs.\ 86.4\%. Charts span 37 countries and the years 1849--2025. This diversity makes EpiCurveBench especially challenging, requiring methods to generalize across highly variable conditions.

\begin{table}[t]
\centering
\caption{Per-chart statistics of EpiCurveBench (n=1{,}000).}
\label{tab:dataset_stats}
\resizebox{\columnwidth}{!}{%
\begin{tabular}{lrrrrr}
\toprule
\textbf{Statistic} & \textbf{Mean} & \textbf{Median} & \textbf{Std} & \textbf{Min} & \textbf{Max} \\
\midrule
Series per chart           & 3.0   & 2.0   & 2.3     & 1 & 14    \\
Points per series          & 245.9 & 52.0  & 516.3   & 5 & 2{,}148 \\
Total datapoints per chart & 699.0 & 156.0 & 1{,}842.1 & 7 & 8{,}592 \\
\bottomrule
\end{tabular}
}
\end{table}

Geographically, the United States dominates (775 charts), reflecting CDC data availability; Canada contributes 28 and Haiti (14) is the most represented lower-income country, with the remaining charts spanning 34 additional countries across Africa, Asia, Europe, and Latin America. By disease, respiratory illnesses dominate---influenza (413), COVID-19 (303), and RSV (165)---with the remainder covering cholera, adenovirus, and others. Full distributions are reported in Appendix~\ref{app:dataset_distributions}.

\section{EpiCurveSimilarity}
\label{sec:metric}
\begin{figure*}[t]
\centering
\includegraphics[width=\textwidth]{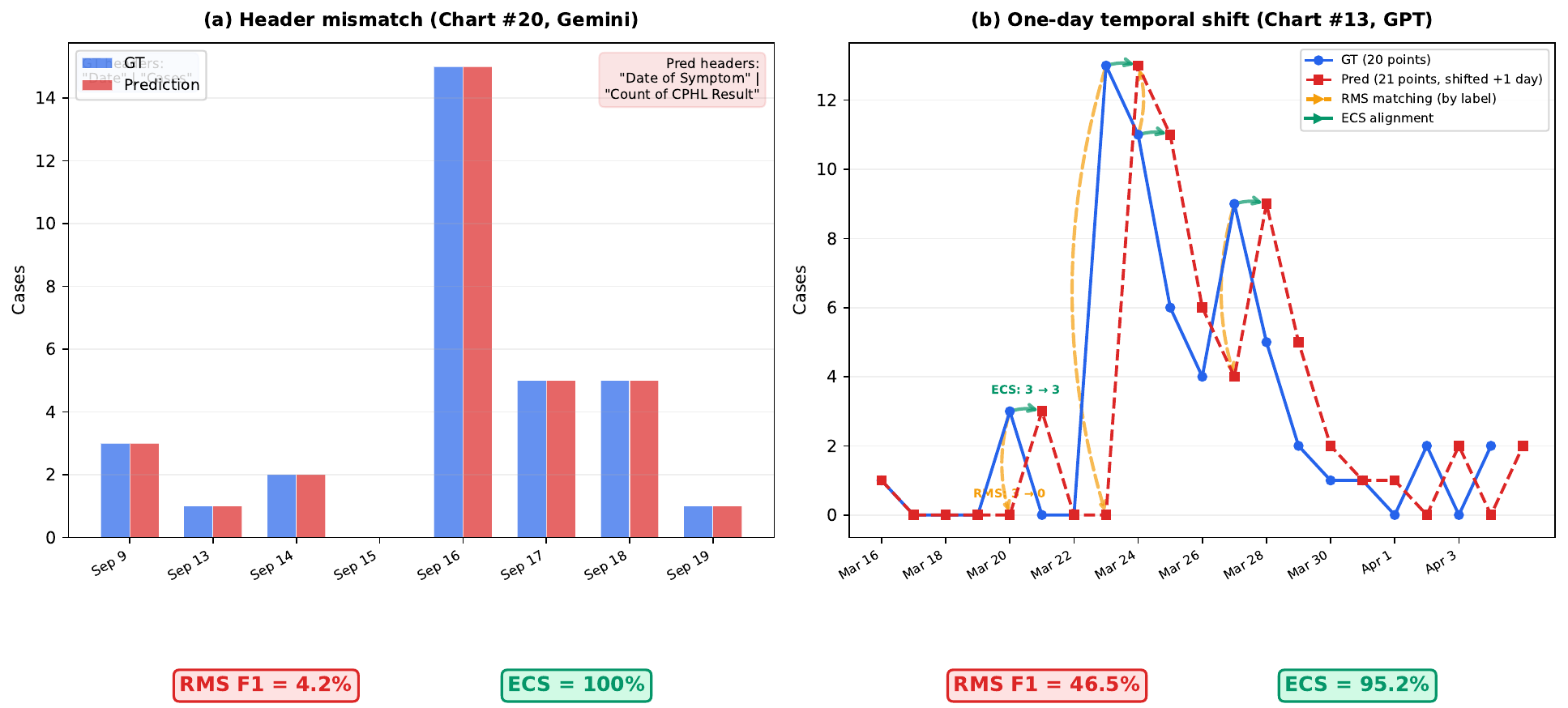}
\caption{Two failure modes of RMS on EpiCurveBench. \textbf{(a) Header mismatch}: Gemini's extraction of Chart \#20 is exact but uses different column headers; RMS scores F1 = 4.2\% vs.\ ECS 100\%. \textbf{(b) One-day temporal shift}: GPT's extraction of Chart \#13 is value-correct but offset by one day; RMS scores F1 = 46.5\% vs.\ ECS 95.2\%.}
\label{fig:metric_comparison}
\end{figure*}

As discussed in Section~\ref{sec:related_work}, RMS and SCRM treat extracted data as unordered key-value mappings and penalize any temporal shift or label mismatch as a complete miss, even though extracted curves frequently exhibit small temporal offsets, boundary gaps, or header differences that preserve the overall trend and remain usable downstream after minimal realignment. A more appropriate evaluation should accommodate local shifts and missing values, penalize them proportionally, and avoid text-based matching of x-axis labels (which are often sparsely or inconsistently formatted). Figure~\ref{fig:metric_comparison} illustrates two such failure modes on real examples: a perfect extraction with mismatched headers (panel a) and a value-correct curve shifted by one day (panel b), both of which RMS scores as near-failures while ECS recognizes as essentially correct.

To address these limitations, we introduce EpiCurveSimilarity (ECS), a metric specifically designed for evaluating extracted time series. ECS builds on Edit Distance with Real Penalty \citep[ERP]{erp}, which uses dynamic programming to align series while accounting for gaps and unmatched regions, and extends it along three axes: \emph{(i)} label-similarity-based series matching, so the metric handles multi-series charts whose extracted and ground-truth series may differ in order or naming; \emph{(ii)} per-series y-axis normalization, so substitution penalties are comparable across series of different scale; and \emph{(iii)} a tolerance threshold that distinguishes near-correct from clearly-wrong matches. Like ERP, ECS supports insertions, deletions, and substitutions, and applies real-valued penalties to quantify differences between matched points.

Since epicurve images often contain multiple time series, we first match each automatically extracted series (extraction methods are introduced in Section \ref{sec:methods}) to its ground-truth, annotated counterpart. Let $E = \{e_1, \dots, e_m\}$ denote the set of extracted series from an image and $T = \{t_1, \dots, t_n\}$ the corresponding set of ground-truth series for that image. Following prior work~\citep{deplot, deplot_cited}, we perform matching using the Normalized Levenshtein Similarity (NLS), defined as $1 - d_L(s_1, s_2)/\max(|s_1|,|s_2|)$ where $d_L$ is the Levenshtein edit distance, which ranges from 0 (no overlap) to 1 (identical labels). Two series are matched if their NLS exceeds 0.5. Extracted series with no matching ground-truth are ignored, while unmatched ground-truth series are assigned a score of zero.

For each matched pair of series $p = (p_1, \dots, p_M)$ and $t = (t_1, \dots, t_N)$, where $i$ and $j$ index values in the predicted and ground-truth series respectively, the ERP distance is computed by evaluating the following recurrence relation for $0\leq i\leq M$ and $0\leq j\leq N$, using dynamic programming:

$$
C(i, j) =
\min \begin{cases}
C(i\!-\!1, j\!-\!1) + D_\theta(p_i, t_j)\\
C(i\!-\!1, j) + \lambda\\
C(i, j\!-\!1) + \lambda
\end{cases}
$$

with boundary conditions $C(0,0)=0$, $C(i,0)=i \cdot \lambda$, and $C(0,j)=j \cdot \lambda$. We fix the gap penalty at $\lambda=1$ (as in~\citet{deplot}). The match cost $D_\theta(p_i,t_j)$ is the relative difference $\delta_{ij} = |p_i - t_j|/(y_{\max}-y_{\min})$, normalized by the series' y-axis range, and clipped to 1 when it exceeds $\theta=0.01$. Thus differences smaller than 1\% of the y-axis range are penalized proportionally; larger differences are treated as entirely incorrect. The EpiCurveSimilarity (ECS) score for a series pair normalizes the alignment cost $C(M,N)$ to $[0,1]$:
$$
\text{ECS}(p, t) = 1 - \frac{C(M, N)}{\#\text{matches} + \#\text{gaps}}.
$$

For each image, we compute the mean ECS across all matched series, assigning a score of zero to any unmatched ground-truth series. The final performance measure is then obtained by averaging these image-level scores across the entire dataset. While we develop ECS in the context of epicurves, its design assumptions---ordered time-indexed series, sparsely labeled x-axes, and tolerance to local temporal shifts---apply to any structured time-series chart-extraction setting, and the gap penalty $\lambda$ and threshold $\theta$ can be tuned for other domains. Appendix~\ref{app:sensitivity} shows that the model ranking and four-model spread reported in Section~\ref{sec:results} are preserved across a wide grid of $(\theta, \lambda, \text{NLS})$ values, with the spread collapsing only in known degenerate regimes ($\theta \to 0$ or $\lambda$ well below average substitution cost).

\section{Experiments}
\subsection{Methods Evaluated}
\label{sec:methods}
We evaluate several methods on EpiCurveBench, including two state-of-the-art specialized chart-extraction models: OneChart~\citep{onechart} and TinyChart~\citep{tinychart}. We also test three frontier VLMs: GPT-5.2~\citep{gpt52}, Claude Opus 4.5~\citep{opus45}, and Gemini 2.5 Pro~\citep{gemini25pro}, as well as a state-of-the-art open VLM (Qwen3-VL-235B-A22B-Instruct~\citep{qwen}). We evaluate the four general-purpose VLMs under three configurations: minimal reasoning effort, high reasoning effort, and high reasoning effort with access to a Python code-execution tool capable of zooming and cropping image regions. All four models use the same generic extraction prompt instructing the model to return a TSV with all timepoints (no gaps), use y-axis or legend labels as headers, and emit \texttt{nan} for missing values. The full prompt text is provided in Appendix~\ref{app:prompt}.

\subsection{Results}
\label{sec:results}
Table~\ref{tab:results} reports ECS scores and inference costs across all evaluated methods and configurations.

\begin{table}[ht]
\centering
\caption{ECS (\%) of all evaluated methods on EpiCurveBench.}
\label{tab:results}
\resizebox{\columnwidth}{!}{%
\begin{tabular}{lcccc}
\toprule
\textbf{Model} & \textbf{Full} & \textbf{Set 1} & \textbf{Set 2} & \textbf{Cost (\$)} \\
\midrule
OneChart            & 9.4  & 8.3  & 18.5 & --- \\
TinyChart           & 11.4 & 10.7 & 18.4 & --- \\
\cmidrule(lr){1-5}
Qwen3-VL Minimal         & 26.1 & 25.2 & 35.2 & 0.36 \\
Qwen3-VL High            & 27.5 & 26.7 & 34.7 & 1.98 \\
Qwen3-VL High+Code       & 29.5 & 28.6 & 37.3 & 4.82 \\
\cmidrule(lr){1-5}
GPT-5.2 Minimal          & 45.4 & 45.1 & 46.6 & 1.71 \\
GPT-5.2 High             & 43.9 & 43.3 & 50.0 & 16.96 \\
GPT-5.2 High+Code        & 44.8 & 44.9 & 44.1 & 19.64 \\
\cmidrule(lr){1-5}
Claude Opus 4.5 Minimal  & 41.3 & 41.6 & 39.4 & 3.92 \\
Claude Opus 4.5 High     & 42.0 & 42.1 & 41.4 & 29.60 \\
Claude Opus 4.5 High+Code & 40.5 & 40.5 & 40.2 & 82.41 \\
\cmidrule(lr){1-5}
Gemini 2.5 Pro Minimal   & 40.5 & 39.4 & 50.2 & 6.29 \\
Gemini 2.5 Pro High      & \textbf{52.3} & \textbf{51.8} & 56.9 & 15.75 \\
Gemini 2.5 Pro High+Code & 48.0 & 45.2 & \textbf{69.4} & 19.16 \\
\bottomrule
\end{tabular}
}
\end{table}

Performance remains far from saturated: even the strongest configuration (Gemini 2.5 Pro at high reasoning effort) reaches only 52.3\% ECS on the full benchmark, underscoring the difficulty of real-world epicurve digitization. The two specialized chart-extraction systems (OneChart 9.4\%, TinyChart 11.4\%) perform poorly, likely reflecting both distribution shift from cleaner, more synthetic training data and reduced model capacity relative to the general-purpose VLMs. Among the four general-purpose VLMs, Gemini 2.5 Pro leads overall, while Qwen3-VL---the only open-weight general-purpose VLM in our evaluation---reaches 27.5\% ECS at high effort for \$1.98 inference cost (29.5\% with code-execution at \$4.82).

\paragraph{Reasoning and tool use have non-uniform effects.} Moving from minimal to high reasoning effort helps Gemini dramatically ($+11.8$ ECS) but yields only modest gains for Qwen3-VL ($+1.4$) and Claude ($+0.7$), and actually \emph{hurts} GPT-5.2 ($-1.5$). Adding Python code-execution on top of high reasoning is equally inconsistent: Qwen3-VL ($+2.0$) and GPT-5.2 ($+0.9$) benefit modestly on the full benchmark, while Claude ($-1.5$) and Gemini ($-4.3$) regress. Tool cost also varies sharply: Claude High+Code costs \$82.41 ($\sim$3$\times$ Claude High), while GPT-5.2 and Gemini incur only modest tool surcharges. Inspecting failures where the tool hurts, we find that by far the dominant failure mode is inaccurate cropping: the model crops overlapping or incomplete regions of the chart, producing TSVs with duplicate or missing rows.

\paragraph{Set 2 vs.\ Set 1.} Scores are higher on Set 2 than Set 1 for all configurations except all three Claude configurations and GPT-5.2 High+Code---counterintuitive given Set 2's curated diversity, but explained by Set 1's heavy tail of very long series (90th-percentile length 1{,}304 vs.\ 224 points) inflating missed-datapoint losses; on numerical error and label mismatch, Set 2 is in fact harder (Section~\ref{sec:error_analysis}).

\paragraph{Chart type and cumulative status.} Stratified results (Appendix~\ref{app:stratified}) show bar charts are uniformly easier than line charts across all models (e.g., Gemini 72.7\% vs.\ 47.5\%), driven by lower datapoint density; Claude and Gemini also score $\sim$20 points higher on cumulative than non-cumulative charts, likely because of both lower complexity and the monotonicity constraint.

\subsection{Error Analysis}
\label{sec:error_analysis}

\begin{figure}[!t]
\centering
\includegraphics[width=\columnwidth]{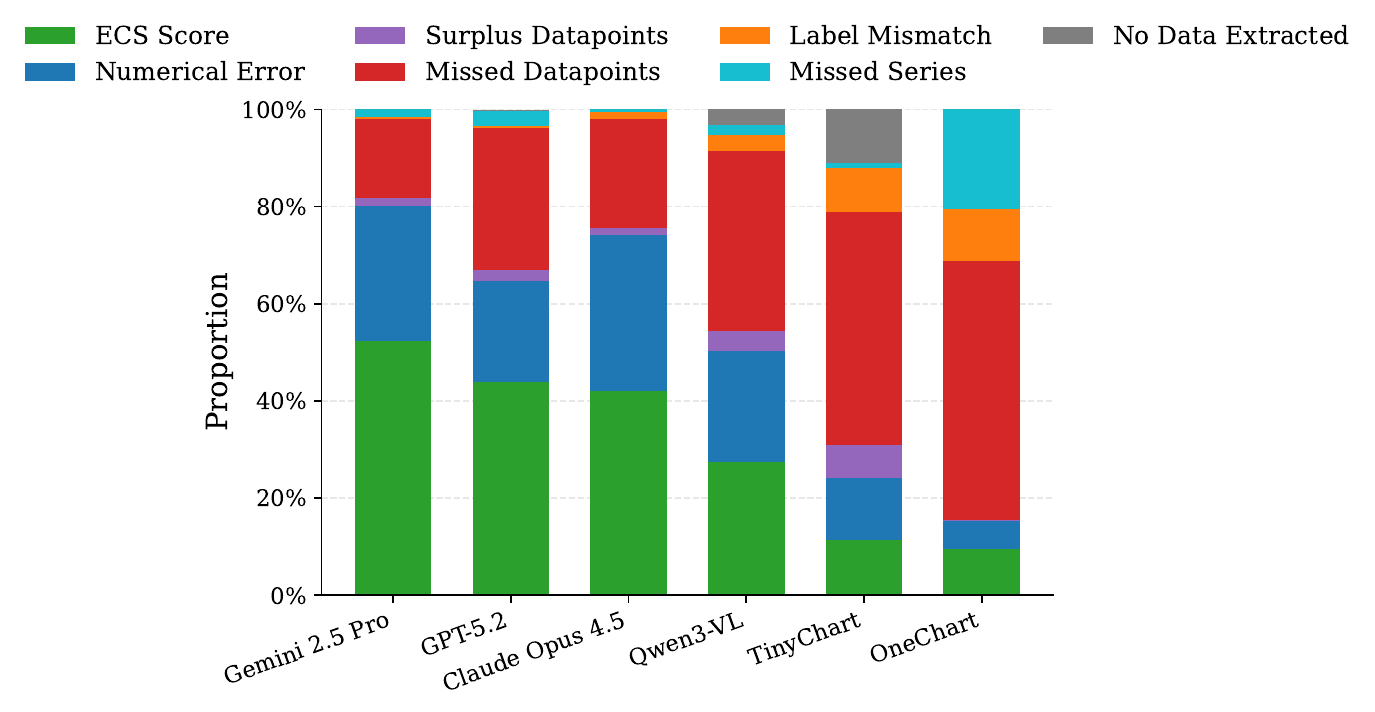}
\caption{Breakdown of ECS scores and error types across models. The green portion represents the achieved ECS score; remaining portions show error contributions. \textit{Numerical Error}: distance between matched points; \textit{Surplus Datapoints}: insertions in the predicted series; \textit{Missed Datapoints}: deletions in the predicted series; \textit{Label Mismatch}: extracted series label does not match any ground-truth label; \textit{Missed Series}: series present in the ground truth but not extracted; \textit{No Data Extracted}: model did not extract any data.}
\label{fig:error_analysis}
\end{figure}

To understand where models fail, we decompose each model's performance into seven components that sum to 100\%. Figure~\ref{fig:error_analysis} shows the breakdown for each model, with the green portion representing the achieved ECS score and the remaining portions representing different types of errors.

For the general-purpose VLMs, numerical errors---imprecise extraction of series values---account for the largest or second-largest share of ECS loss (20--30\%), reflecting the well-documented difficulty multimodal models have reading fine-grained visual elements such as bar heights~\citep{amplified,plottwist}. Missed datapoints are also a top-two contributor for every model, and dominate for the specialized models OneChart and TinyChart (over half of total loss), often because they extract at a coarser temporal resolution than the chart contains (e.g., weekly instead of daily); the specialized models also fail more frequently on series-label matching.

\paragraph{Set 1 vs.\ Set 2.} To substantiate the claim in Section~\ref{sec:results} that Set 2's higher aggregate ECS does not mean it is uniformly easier, we decompose the three largest error components by set for the two strongest general-purpose VLMs (Table~\ref{tab:error_by_set}, Appendix~\ref{app:error_by_set}). Set 1's heavy tail of very long series inflates missed-datapoint losses (e.g., Gemini 17.5\% on Set 1 vs.\ 5.4\% on Set 2; GPT-5.2 31.1\% vs.\ 12.3\%), but on numerical error and label mismatch Set 2 is in fact harder, consistent with its greater stylistic diversity. The aggregate ``easier'' result for Set 2 is therefore driven entirely by the missed-datapoints gap, not by an across-the-board reduction in difficulty.

\subsection{Metric Comparison: ECS vs.\ RMS, SCRM, and DTW}
\label{sec:metric_comparison}

To validate the utility of ECS, we compare it against three alternative chart-extraction metrics: \textbf{RMS}~\citep{deplot} and \textbf{SCRM}~\citep{scrm}, which score extractions as unordered key-value cell mappings, and \textbf{DTW}~\citep{dtw}, which aligns the two series via many-to-one warping but without a gap penalty for unmatched points. SCRM is instantiated with the chart-type-aware numeric tolerances suggested by~\citet{scrm}: 5\% for bar charts and 10\% for line and mixed charts. DTW uses the same per-series y-axis normalization and tolerance-clipping rule as ECS. Table~\ref{tab:metric_comparison} reports all four metrics on the full benchmark (n=1{,}000).

\begin{table}[ht]
\centering
\caption{Comparison of ECS against three alternative metrics on EpiCurveBench (n=1{,}000). For the four general-purpose VLMs, scores are at high reasoning effort (the ``High'' rows of Table~\ref{tab:results}); OneChart and TinyChart have a single configuration. The bottom row reports the spread (max\,$-$\,min) across the four general-purpose VLMs for each metric.}
\label{tab:metric_comparison}
\resizebox{\columnwidth}{!}{%
\begin{tabular}{lcccc}
\toprule
\textbf{Model} & \textbf{RMS (\%)} & \textbf{SCRM (\%)} & \textbf{DTW (\%)} & \textbf{ECS (\%)} \\
\midrule
OneChart        & 4.4  & 4.3  & 26.4 & 9.4  \\
TinyChart       & 8.5  & 8.5  & 35.6 & 11.4 \\
Qwen3-VL        & 18.0 & 17.9 & 52.2 & 27.5 \\
GPT-5.2         & 18.6 & 18.6 & 65.8 & 43.9 \\
Claude Opus 4.5 & 22.8 & 23.0 & 62.4 & 42.0 \\
Gemini 2.5 Pro  & 20.5 & 20.8 & 72.9 & 52.3 \\
\midrule
\textit{4-VLM spread} & 4.8 & 5.2 & 20.7 & \textbf{24.8} \\
\bottomrule
\end{tabular}
}
\end{table}

The four metrics split into two regimes. The two \emph{key-value} metrics (RMS and SCRM) compress the four general-purpose VLMs into a \mbox{$\sim$5-point} band (RMS spread 4.8, SCRM spread 5.2) and agree that Claude Opus 4.5 is the top performer. The two \emph{temporal-alignment} metrics (DTW and ECS) instead spread the four VLMs over a 20--25-point range (DTW spread 20.7, ECS spread 24.8) and agree that Gemini 2.5 Pro is the top performer.

\textbf{RMS and SCRM both fail to discriminate}, despite SCRM's chart-type-aware tolerances. As discussed in Section~\ref{sec:metric}, this is because both metrics treat extracted points as unordered key-value mappings, penalizing any temporal shift as a complete mismatch. SCRM's relaxed numeric tolerance reduces a few near-correct values' penalties but does not change the dominant failure mode (Figure~\ref{fig:metric_comparison}): when an extracted curve is offset by a few time steps, both RMS and SCRM assign near-zero credit even if the values themselves are accurate. Since the general-purpose VLMs vary substantially in the degree to which their errors are dominated by such temporal misalignments, both key-value metrics collapse them into a narrow band and identify the \emph{wrong} top performer (Claude rather than Gemini).

\textbf{DTW and ECS both discriminate well} because both align the two series along a monotone path before scoring values, recognizing partial temporal correspondence rather than penalizing it catastrophically. DTW achieves a similar four-VLM spread to ECS (20.7 vs.\ 24.8 points) and the same model ranking among the general-purpose VLMs, suggesting that the bulk of the discrimination gain over RMS/SCRM comes from \emph{any} temporal-alignment mechanism. The advantage of ECS over DTW is therefore in \emph{downstream meaningfulness}, which we evaluate next.

\subsection{Downstream Relevance}
\label{sec:downstream}
A meaningful evaluation metric should correlate with downstream task performance: better extraction quality should translate to more accurate derived quantities. To validate ECS---and to test whether ECS's gap penalty actually buys downstream-relevant information that the gap-free DTW lacks (Section~\ref{sec:metric_comparison})---we examine whether higher extraction-quality scores correspond to lower errors in computing four downstream summary statistics: total case/death counts, peak timing, peak magnitude, and growth rate. The first measures aggregate burden, the second and third locate and quantify the epidemic peak, and the fourth captures early-phase exponential dynamics---all of which are routine inputs to epidemic forecasting.

Each statistic is computed per matched series across all four general-purpose VLMs, with statistic-specific filters that account for what each quantity requires to be well-defined. For \textbf{total counts}, we restrict to series whose labels indicate case or death counts (excluding rates, percentages, and ratios), since relative error in aggregate counts is only meaningful for those quantities, and compute the per-series relative error $|S_{\text{pred}} - S_{\text{gt}}| / S_{\text{gt}}$ where $S_{\text{pred}}$ and $S_{\text{gt}}$ are the sums of predicted and ground-truth values; this yields 944 series. For \textbf{peak timing}, \textbf{peak magnitude}, and \textbf{growth rate}---which capture the \emph{shape} of the curve rather than its absolute level, and are therefore meaningful on rates and percentages as well---we retain all matched series with non-zero ground-truth sum, then apply per-statistic well-definedness filters (a defined peak in the ground truth for peak timing, non-zero peak in both prediction and ground truth for peak magnitude, and an ascending phase of sufficient length for growth rate), yielding 10{,}512 / 7{,}484 / 4{,}862 series respectively. Peak timing is the absolute difference between the predicted and ground-truth peak indices; peak magnitude is the relative error of the peak value; and growth rate is the Pearson correlation between predicted and ground-truth log-incidence trajectories over the ascending phase, defined as the segment from the first non-zero ground-truth datapoint to the ground-truth peak index. Table~\ref{tab:downstream} reports the Spearman rank correlation between each of ECS and DTW and each downstream quantity.

\begin{table}[t]
\centering
\caption{Spearman rank correlation between extraction-quality metrics (ECS, DTW) and downstream summary statistics. $n$ counts (series, model) pairs pooled across the four general-purpose VLMs, not unique series. All correlations are in the expected direction and highly significant ($p < 0.001$ in each row). \textbf{ECS correlates 1.5--3.6$\times$ more strongly than DTW with every statistic}, because DTW lacks a gap penalty and so cannot distinguish a truncated prediction from a temporally faithful one.}
\label{tab:downstream}
\begin{tabular}{lrrr}
\toprule
\textbf{Statistic} & $\boldsymbol{r}_{\text{ECS}}$ & $\boldsymbol{r}_{\text{DTW}}$ & $\boldsymbol{n}$ \\
\midrule
Total-count & $-0.58$ & $-0.38$ & 944 \\
Peak timing          & $-0.51$ & $-0.15$ & 10{,}512 \\
Peak magnitude       & $-0.29$ & $-0.08$ & 7{,}484 \\
Growth-rate    & $+0.19$ & $+0.08$ & 4{,}862 \\
\bottomrule
\end{tabular}
\end{table}

All eight correlations are in the expected direction and highly significant. ECS correlates substantially more strongly than DTW with every downstream statistic: 1.5$\times$ stronger on total-count error, 3.4$\times$ on peak timing, 3.6$\times$ on peak magnitude, and 2.3$\times$ on growth-rate fidelity. DTW allows arbitrary many-to-one warping with no gap penalty, so a prediction that truncates after a few datapoints can freely warp to align with whatever ground-truth points it has, yielding a misleadingly high similarity score even when most of the epidemic's temporal shape is missing. ECS, by contrast, charges a penalty for each insertion or deletion in the alignment, so a truncated extraction loses score in direct proportion to its missing datapoints---which is the failure mode that propagates into wrong peak times, wrong peak magnitudes, and wrong totals downstream. The consistent ECS advantage across total-count error, peak timing, peak magnitude, and growth-rate fidelity provides additional evidence that ECS rewards extractions that preserve both magnitude and temporal shape of the epidemic.

\section{Conclusion}
We contribute EpiCurveBench, a benchmark of 1{,}000 real-world epicurves on which the strongest of six evaluated VLMs reaches only 52.3\% ECS, and EpiCurveSimilarity (ECS), a metric that aligns predicted and ground-truth series via dynamic programming and an explicit gap penalty. ECS spreads the four general-purpose VLMs across a 25-point range where key-value metrics (RMS, SCRM) compress them into $\sim$5 points, and correlates 1.5--3.6$\times$ more strongly than DTW with downstream errors in total counts, peak timing, peak magnitude, and growth-rate fidelity. Benchmark and metric apply to any structured time-series chart-extraction setting, and the diagnostic error breakdown points to concrete weaknesses---most prominently numerical imprecision on dense series and truncation of long series---for future models to address.

\section*{Limitations}
EpiCurveBench has several limitations. First, although the benchmark aggregates epicurves from diverse sources, countries, and time periods, \textbf{Set 1 is currently dominated by charts from the United States}. This reflects the rarity of openly licensed epicurves paired with machine-readable ground truth, and the fact that the CDC is one of the few organizations that routinely publishes both figures and underlying data. This geographic skew may bias evaluation toward the visual conventions, labeling practices, and reporting formats used by the CDC. A complexity-matched analysis (Appendix~\ref{app:cdc_stratification}) finds that the apparent CDC vs.\ non-CDC accuracy gap is driven by series-length differences rather than CDC-specific overfitting, but expanding coverage of underrepresented regions and institutions remains important.

Second, the benchmark exhibits \textbf{domain and format skews} driven by what is available with ground truth. In particular, EpiCurveBench is overrepresented by respiratory illnesses, line charts, and non-cumulative time series. Future dataset expansions should explicitly target underrepresented strata.

Finally, although we focus on epicurves, our framework may extend to \textbf{other types of charts} where extracting structured data could unlock valuable evidence, such as histograms of chronic disease prevalence, laboratory test positivity trends, and biometric or environmental time series. Extending EpiCurveBench-style evaluation to these settings would help assess whether progress on epicurve digitization transfers to the broader landscape of visualizations.

\bibliography{custom}

\appendix

\section{Extraction Prompt}
\label{app:prompt}

All four general-purpose VLMs were evaluated with the following extraction prompt:

\begin{quote}\small
Here is an image of a chart.

Please extract the numerical data it represents and return it in TSV (tab-separated values) format with appropriate headers.

For time-series charts, extract datapoints for **all timepoints without any gaps**, even if a timepoint is **not explicitly shown on the x axis**.

For series column headers, **use the y axis label if there is a single series in the image, and the series label (e.g., from the legend) if there is more than one series**.

When a value is missing, **use "nan"**.

Copy the headers exactly as they are in the image where applicable.

IMPORTANT: For the TSV, use a tab character (\texttt{\textbackslash t}) as the separator.

Remember: The sole output should be the TSV table surrounded by ```tsv``` and nothing else.
\end{quote}

\section{Inter-Annotator Agreement Stratification}
\label{app:annotation_agreement}

Stratifying the inter-annotator ECS reported in Section~\ref{sec:dataset_characteristics} by chart type, agreement was 90.6\% on bar charts, 89.5\% on line charts, and 92.2\% on charts containing both. Stratifying by series length, agreement was 98.7\% on short series ($\le 20$ points) and 81.7\% on long series ($>100$ points), indicating that disagreement grows with series density.

\section{Per-Set Error Decomposition}
\label{app:error_by_set}

Table~\ref{tab:error_by_set} decomposes the three largest error components by set (Set 1 vs.\ Set 2) for the two strongest general-purpose VLMs, supporting the analysis in Sections~\ref{sec:error_analysis} and~\ref{app:cdc_stratification}.

\begin{table}[h]
\centering
\caption{Per-set decomposition of the three largest error components for Gemini 2.5 Pro and GPT-5.2 at high reasoning effort. Values are fractions of total possible ECS (e.g., 0.175 means missed datapoints account for 17.5\% of ECS loss). Bold cells mark the harder set for each component.}
\label{tab:error_by_set}
\begin{tabular}{lcccc}
\toprule
 & \multicolumn{2}{c}{\textbf{Gemini}} & \multicolumn{2}{c}{\textbf{GPT}} \\
\cmidrule(lr){2-3}\cmidrule(lr){4-5}
\textbf{Component} & Set 1 & Set 2 & Set 1 & Set 2 \\
\midrule
Missed datapoints & \textbf{0.175} & 0.054 & \textbf{0.311} & 0.123 \\
Numerical error   & 0.275 & \textbf{0.301} & 0.197 & \textbf{0.299} \\
Label mismatch    & 0.000 & \textbf{0.047} & 0.005 & \textbf{0.012} \\
\bottomrule
\end{tabular}
\end{table}

\section{Source, Country, and Disease Distribution}
\label{app:dataset_distributions}

\begin{figure}[h]
\centering
\includegraphics[width=\columnwidth]{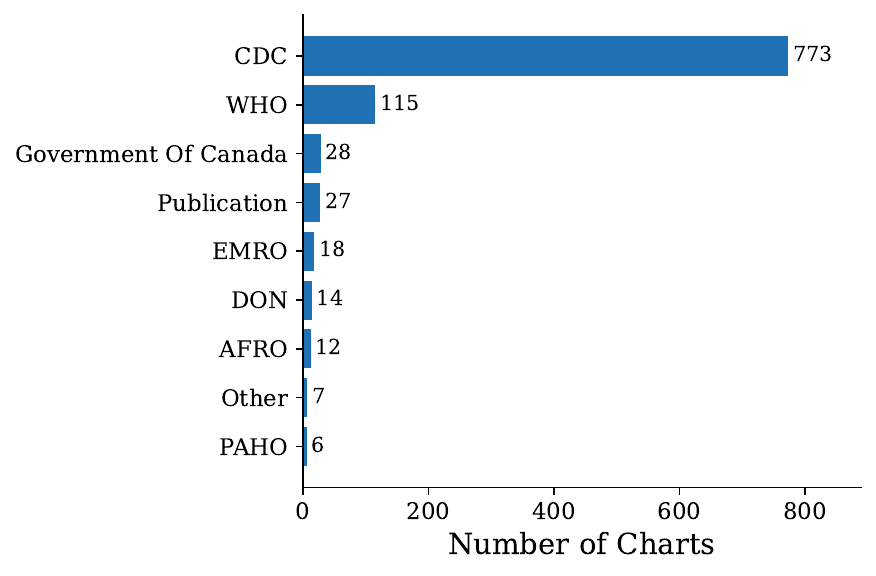}
\caption{Distribution of EpiCurveBench sources by origin. There are 13 unique source types. The ``Other'' category includes sources that appear only once. Most publication sources come from ScienceDirect and ResearchGate.}
\label{fig:epicurve_sources_distribution}
\end{figure}

\begin{figure}[h]
\centering
\includegraphics[width=\columnwidth]{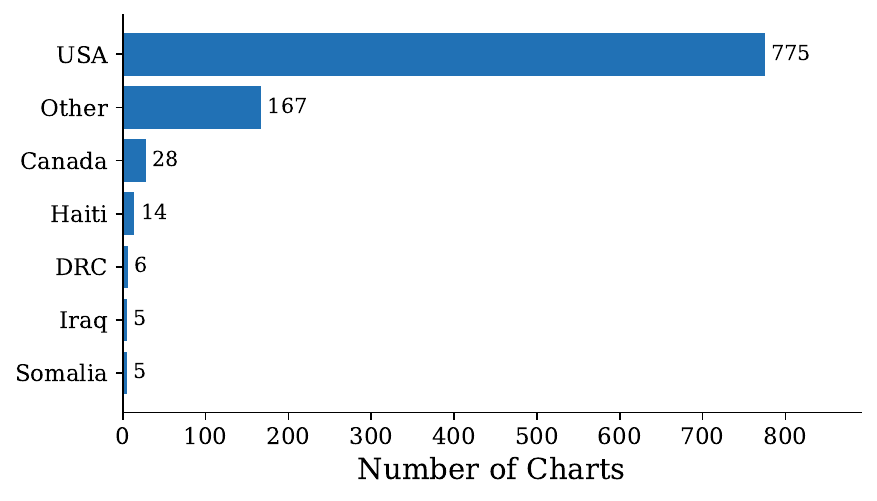}
\caption{Geographic distribution of charts in EpiCurveBench.}
\label{fig:country_distribution}
\end{figure}

\begin{figure}[h]
\centering
\includegraphics[width=\columnwidth]{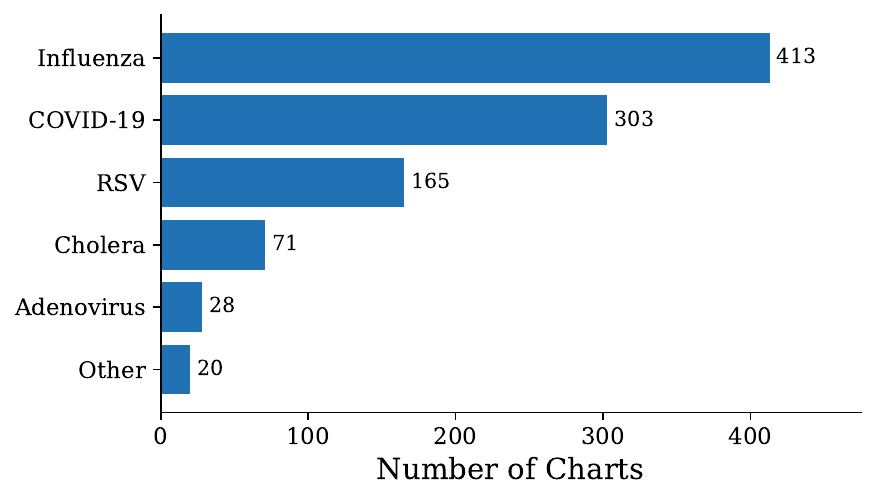}
\caption{Distribution of diseases represented in EpiCurveBench.}
\label{fig:disease_distribution}
\end{figure}

Figure~\ref{fig:epicurve_sources_distribution} shows the distribution of sources. Figure~\ref{fig:country_distribution} shows the geographic distribution of charts. Figure~\ref{fig:disease_distribution} shows the distribution of diseases.

\section{Performance by Chart Type and Cumulative Status}
\label{app:stratified}

\begin{figure}[h]
\centering
\includegraphics[width=\columnwidth]{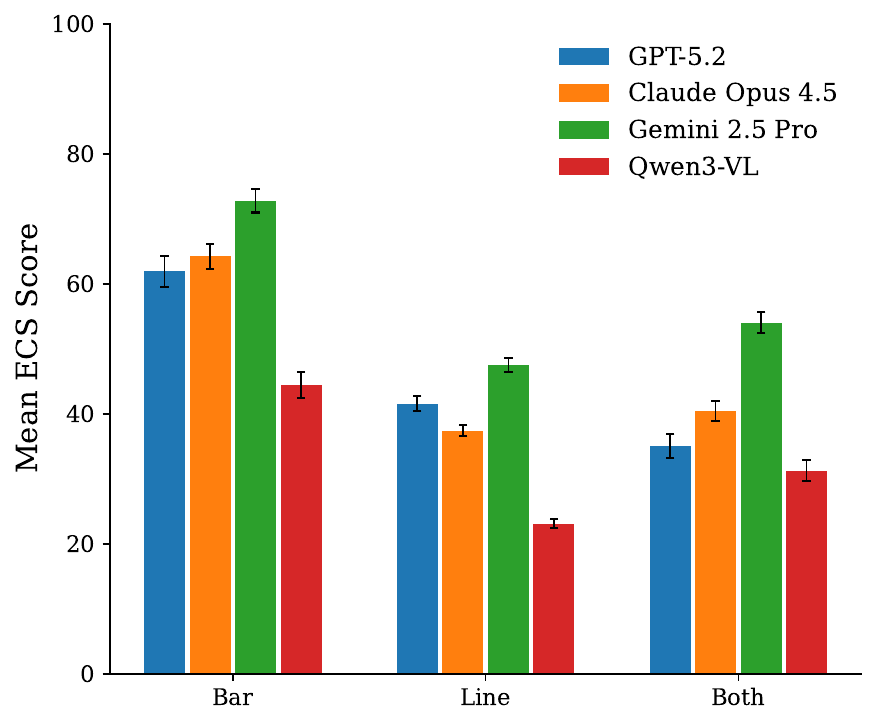}
\caption{Mean ECS score by chart type.}
\label{fig:ecs_by_chart_type}
\end{figure}

\begin{figure}[h]
\centering
\includegraphics[width=\columnwidth]{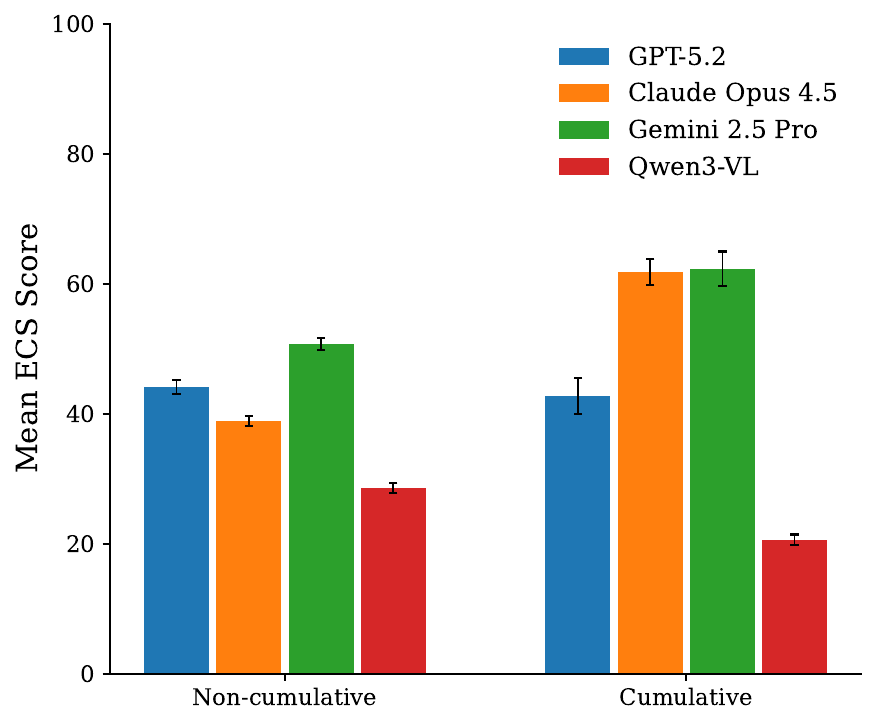}
\caption{Mean ECS score by cumulative status.}
\label{fig:ecs_by_cumulative}
\end{figure}

We stratify performance by chart type (bar, line, or both) and by whether the chart displays cumulative or non-cumulative case counts. Figures~\ref{fig:ecs_by_chart_type} and~\ref{fig:ecs_by_cumulative} show the results for the four general-purpose VLMs.

\textbf{Chart type.} Bar charts are consistently the easiest to extract across all models: Gemini 2.5 Pro achieves 72.7\% ECS on bar charts, compared to 47.5\% on line charts and 54.0\% on charts containing both. This pattern holds for all evaluated models. The difficulty of line charts can be attributed to dataset characteristics: bar charts in EpiCurveBench have an average of 345 datapoints, whereas line charts average 842 datapoints. The higher datapoint density of line charts makes precise extraction more challenging.

\textbf{Cumulative status.} Performance on cumulative charts is notably higher for Claude Opus 4.5 and Gemini 2.5 Pro, both achieving approximately 62\% ECS on cumulative charts versus 39--51\% on non-cumulative charts. Cumulative charts in the dataset also tend to be simpler, with an average of 99 datapoints, compared to 787 datapoints for non-cumulative charts. The monotonically non-decreasing nature of cumulative data may also provide an advantage, as models can leverage the constraint that each value must be at least as large as its predecessor. Interestingly, Qwen3-VL shows the opposite pattern, performing worse on cumulative charts (21\%) than non-cumulative charts (29\%), suggesting model-specific sensitivities to chart characteristics.

\section{Source Stratification: CDC vs.\ Non-CDC}
\label{app:cdc_stratification}

Because the CDC accounts for the majority of Set 1, a natural concern is that models may simply have memorized CDC visual conventions rather than learned general epicurve understanding. The raw CDC vs.\ non-CDC ECS gap appears large at first glance (e.g., Gemini 2.5 Pro scores 58.7\% on CDC charts vs.\ 30.4\% on non-CDC charts), but the two strata are not comparable: non-CDC charts have median series length 242 vs.\ 52 for CDC, so the gap is confounded by chart complexity. To control for this, we restrict to series of length $\le 100$ (the regime where CDC and non-CDC overlap) and report the resulting per-model ECS in Table~\ref{tab:cdc_stratification}.

\begin{table}[h]
\centering
\caption{ECS (\%) on charts with series length $\le 100$, stratified by source. Controlling for series complexity, the raw CDC advantage disappears or reverses for three of the four general-purpose VLMs.}
\label{tab:cdc_stratification}
\begin{tabular}{lccc}
\toprule
\textbf{Model} & \textbf{CDC} & \textbf{Non-CDC} & $\boldsymbol{\Delta}$ \\
\midrule
Gemini & 60.6 & 64.5 & $-3.9$ \\
Claude & 50.0 & 46.0 & $+4.1$ \\
GPT-5.2 & 51.6 & 60.6 & $-9.1$ \\
Qwen & 31.8 & 42.3 & $-10.5$ \\
\bottomrule
\end{tabular}
\end{table}

After complexity matching, the CDC advantage shrinks substantially and reverses for three of the four general-purpose VLMs, which in fact perform \emph{better} on non-CDC charts of equivalent length; only Claude Opus 4.5 retains a small residual CDC advantage ($+4.1$ ECS). This suggests that the raw CDC vs.\ non-CDC gap is driven primarily by series length, not by overfitting to CDC-specific visual conventions, and that the headline ECS scores generalize beyond the CDC subset of the benchmark.

The complexity-matched analysis above intentionally restricts to series length $\le 100$, the regime in which CDC and non-CDC charts overlap. To address whether CDC overfitting could still drive the gap at longer series lengths, we appeal to the per-set error decomposition reported in Section~\ref{sec:error_analysis} (Table~\ref{tab:error_by_set}). The error components diagnostic of style memorization---numerical error and label mismatch---are in fact \emph{higher} on the source-diverse Set 2 (13 source types) than on the CDC-dominated Set 1 (85\% CDC): numerical error rises from 0.275 to 0.301 for Gemini and from 0.197 to 0.299 for GPT-5.2, and label mismatch from 0.000 to 0.047 (Gemini) and 0.005 to 0.012 (GPT-5.2). The only component that moves in the opposite direction is missed datapoints, which is a series-length / truncation artifact rather than a style artifact. If models had memorized CDC visual conventions, we would expect numerical and label-matching errors to be lower on the CDC-dominated set, not higher. The remaining long-series CDC gap is therefore attributable to truncation losses on long inputs rather than to CDC-specific style.

\section{Hyperparameter Sensitivity}
\label{app:sensitivity}

ECS has three consequential hyperparameters: the tolerance threshold $\theta$ (default 0.01), the gap penalty $\lambda$ (default 1.0), and the NLS series-matching threshold (default 0.5). To verify that the model ranking and the 25-point spread reported in Section~\ref{sec:results} are not artifacts of these choices, we recompute ECS for the four general-purpose VLMs (high reasoning effort) while sweeping each hyperparameter and holding the other two at their defaults.

Tables~\ref{tab:sens_theta}--\ref{tab:sens_text_theta} report per-model ECS on the full dataset, the four-model spread, and the resulting ranking at each grid point. The ranking \textit{Gemini~$>$~GPT~$>$~Claude~$>$~Qwen} is preserved in 15 of the 16 configurations; the sole exception is the strictest tolerance, $\theta = 0.001$, where Claude (29.4) edges past GPT (28.4) by 1.0 ECS---a near-tie in a regime where almost every prediction is judged incorrect, compressing scores at the floor. The four-model spread stays in the 22--28-point range across the $\theta \ge 0.005$ portion of the $\theta$ sweep, all of the NLS sweep, and the $\lambda \ge 1$ portion of the $\lambda$ sweep, and collapses only at very low $\lambda$ (where the optimal alignment prefers cheap gaps to substitutions, so ECS saturates near $1 - \lambda$ for every model regardless of extraction quality). Both the floor regime at $\theta = 0.001$ and the low-$\lambda$ collapse are explained by the metric's known structural behavior rather than by ranking instability, and lie outside any reasonable choice of hyperparameter. The NLS threshold has the smallest effect: every model gains a few points monotonically as the threshold tightens, but the ranking and spread are essentially unchanged across the full $[0, 0.9]$ range. Overall, the 25-point spread reported in the main results is a stable property of the metric across a wide range of hyperparameter settings, not an artifact of the specific choice $(\theta, \lambda, \text{NLS}) = (0.01, 1.0, 0.5)$.

\begin{table}[h]
\centering
\caption{ECS (\%) as the tolerance threshold $\theta$ is varied with $\lambda{=}1$, $\text{NLS}{=}0.5$. Default $\theta{=}0.01$ in bold; \(\Delta\) is the spread between best and worst model.}
\label{tab:sens_theta}
\begin{tabular}{lccccc}
\toprule
$\boldsymbol{\theta}$ & \textbf{Qwen} & \textbf{Claude} & \textbf{GPT} & \textbf{Gemini} & $\boldsymbol{\Delta}$ \\
\midrule
0.001 & 18.0 & 29.4 & 28.4 & 32.1 & 14.1 \\
0.005 & 22.6 & 35.8 & 37.6 & 44.3 & 21.7 \\
\textbf{0.01}  & \textbf{27.5} & \textbf{42.0} & \textbf{43.9} & \textbf{52.3} & \textbf{24.8} \\
0.02  & 34.4 & 50.8 & 51.6 & 62.1 & 27.7 \\
0.05  & 42.5 & 58.6 & 59.8 & 70.6 & 28.1 \\
0.1   & 48.1 & 63.9 & 64.9 & 75.7 & 27.6 \\
\bottomrule
\end{tabular}
\end{table}

\begin{table}[h]
\centering
\caption{ECS (\%) as the gap penalty $\lambda$ is varied with $\theta{=}0.01$, $\text{NLS}{=}0.5$. Default $\lambda{=}1.0$ in bold.}
\label{tab:sens_lambda}
\begin{tabular}{lccccc}
\toprule
$\boldsymbol{\lambda}$ & \textbf{Qwen} & \textbf{Claude} & \textbf{GPT} & \textbf{Gemini} & $\boldsymbol{\Delta}$ \\
\midrule
0.25 & 63.7 & 78.6 & 80.4 & 81.2 & 17.9 \\
0.5  & 39.7 & 55.1 & 55.6 & 61.5 & 21.8 \\
\textbf{1.0}  & \textbf{27.5} & \textbf{42.0} & \textbf{43.9} & \textbf{52.3} & \textbf{24.8} \\
2.0  & 22.5 & 36.1 & 40.9 & 47.9 & 25.4 \\
4.0  & 19.9 & 32.8 & 37.9 & 45.0 & 25.1 \\
\bottomrule
\end{tabular}
\end{table}

\begin{table}[h]
\centering
\caption{ECS (\%) as the NLS series-matching threshold is varied with $\theta{=}0.01$, $\lambda{=}1$. Default 0.5 in bold.}
\label{tab:sens_text_theta}
\begin{tabular}{lccccc}
\toprule
\textbf{NLS} & \textbf{Qwen} & \textbf{Claude} & \textbf{GPT} & \textbf{Gemini} & $\boldsymbol{\Delta}$ \\
\midrule
0.0  & 26.1 & 40.3 & 40.5 & 50.8 & 24.7 \\
0.3  & 27.2 & 41.3 & 43.2 & 52.1 & 24.9 \\
\textbf{0.5}  & \textbf{27.5} & \textbf{42.0} & \textbf{43.9} & \textbf{52.3} & \textbf{24.8} \\
0.7  & 29.2 & 44.4 & 45.0 & 55.2 & 26.0 \\
0.9  & 30.3 & 44.8 & 45.4 & 56.0 & 25.7 \\
\bottomrule
\end{tabular}
\end{table}

\section{Benchmark Comparison: EpiCurveBench vs.\ ChartQA}
\label{app:benchmark_comparison}

To demonstrate that EpiCurveBench provides a more challenging testbed than existing benchmarks, we compare model performance on EpiCurveBench against ChartQA~\citep{chartqa}, a widely used benchmark for chart data extraction. Table~\ref{tab:benchmark_comparison} reports RMS scores on both benchmarks.

\begin{table}[h]
\centering
\caption{RMS performance comparison on EpiCurveBench vs.\ ChartQA. For the four general-purpose VLMs, EpiCurveBench scores are at high reasoning effort (the ``High'' rows of Table~\ref{tab:results}); OneChart and TinyChart have a single configuration.}
\label{tab:benchmark_comparison}
\begin{tabular}{lcc}
\toprule
\textbf{Model} & \textbf{EpiCurveBench} & \textbf{ChartQA} \\
               & RMS (\%) & RMS (\%) \\
\midrule
OneChart        & 4.4  & 35.9 \\
TinyChart       & 8.5  & 95.2 \\
Qwen3-VL        & 18.0 & 89.8 \\
GPT-5.2         & 18.6 & 87.3 \\
Claude Opus 4.5 & 22.8 & 87.7 \\
Gemini 2.5 Pro  & 20.5 & 88.2 \\
\bottomrule
\end{tabular}
\end{table}

On ChartQA, all four general-purpose VLMs achieve RMS scores between 87--90\%, with TinyChart---a specialized chart extraction model---reaching 95.2\%. These near-ceiling scores indicate that ChartQA is approaching saturation: there is little room for further improvement, and the benchmark no longer discriminates between strong models.

On EpiCurveBench, the same models score between 18--23\% RMS, representing a drop of 65--72 percentage points. Even the best-performing model (Claude Opus 4.5 at 22.8\%) leaves substantial room for improvement. This gap reflects the fundamental differences between the benchmarks: ChartQA contains relatively simple charts with few data points and clearly printed values, whereas EpiCurveBench features dense time series with hundreds to thousands of datapoints, overlapping series, and diverse visual styles from real-world public health sources.

\end{document}